\documentclass[conference]{IEEEtran}
\IEEEoverridecommandlockouts
\usepackage{cite}
\usepackage{amsmath,amssymb,amsfonts}
\usepackage{algorithmic}
\usepackage{graphicx}
\usepackage{textcomp}
\usepackage{xcolor}
\usepackage{color, colortbl}
\usepackage{hyperref}

\hypersetup{
    colorlinks=true,
    linkcolor=blue,
    urlcolor=cyan,
    pdfpagemode=FullScreen,
    pdftitle={Self-supervised Learning as a Means to Reduce the Need for Labeled Data in Medical Image Analysis}
    }
\begin{document}

\title{Self-supervised Learning as a Means to Reduce the Need for Labeled Data in Medical Image Analysis\\
\thanks{This work was supported in part by the Croatian Science Foundation under Project UIP-2017-05-4968.}
}

\author{\IEEEauthorblockN{Marin Benčević\IEEEauthorrefmark{1}\IEEEauthorrefmark{2}, Marija Habijan\IEEEauthorrefmark{1}, Irena Galić\IEEEauthorrefmark{1}, Aleksandra Pizurica\IEEEauthorrefmark{3}} 
\IEEEauthorblockA{\IEEEauthorrefmark{1}Faculty of Electrical Engineering, Computer Science and Information Technology\\
J. J. Strossmayer University of Osijek, Osijek, Croatia}
\IEEEauthorblockA{\IEEEauthorrefmark{2}Email: marin.bencevic@ferit.hr} 
\IEEEauthorblockA{\IEEEauthorrefmark{3}TELIN-GAIM, Faculty of Engineering and Architecture\\
Ghent University, Ghent, Belgium}}

\maketitle

\begin{abstract}
\end{abstract}
One of the largest problems in medical image processing is the lack of annotated data. Labeling 
medical images often requires highly trained experts and can be a time-consuming process. In this 
paper, we evaluate a method of reducing the need for labeled data in medical image object detection 
by using self-supervised neural network pretraining. We use a dataset of chest X-ray 
images with bounding box labels for 13 different classes of anomalies. The networks are pretrained 
on a percentage of the dataset without labels and then fine-tuned on the rest of the dataset. We 
show that it is possible to achieve similar performance to a fully supervised model in terms of mean average 
precision and accuracy with only 60\% of the labeled data. We also show 
that it is possible to increase the maximum performance of a fully-supervised model by adding a 
self-supervised pretraining step, and this effect can be observed with even a small amount of 
unlabeled data for pretraining.


\section{Introduction}

In medical image processing, the lack of annotated data is a common obstacle for deep learning models. To function 
robustly and to show their generalizability potential, deep neural networks require a large 
amount of annotated images \cite{litjensSurveyDeepLearning2017}. However, 
annotating medical images often requires time-consuming, tedious work from trained clinitians. 
Hence, there is a huge need to improve the 
data efficiency and robustness of deep learning networks for medical image processing
trained on smaller datasets. Furthermore, 
there is a need to make the labeling process faster to save experts' time. This paper presents a 
step towards both of these goals. The primary contribution of the paper is the evaluation of a method 
that uses self-supervised learning 
to extract salient information from \textit{unlabeled} images which can then be used to more 
easily train an object detection model on a more limited dataset 
of \textit{labeled} images.

We show that it is possible to improve the performance of an object detection model
for medical images by utilizing self-supervised learning in a pretraining phase. In addition, we
also show that it is possible to achieve similar performance to a fully supervised model with
only 60\% of the labeled data. The code for all of the
experiments in this paper is available at \href{https://github.com/marinbenc/ssl-for-medical-object-detection}{github.com/marinbenc/ssl-for-medical-object-detection}.

\subsection{Related work}

There are very few papers published that present methods for object detection in 
chest X-ray images. One notable paper similar to our work is the one by Luo et al.
\cite{luoIntelligentSolutionsChest2021}. They use the same dataset as the experiments
presented in this paper and train a YOLOv5 model with a ResNet50-based backbone.
They achieve results similar to the ones in our experiments,
however, they do not employ any kind of unsupervised learning and train on the full labeled dataset. 
The goal of this paper
is not to beat the state of the art performance, but rather to show if similar results could
be achieved with much less labeled data.

\subsubsection{Improving data efficiency of neural nets}

There are several ways to try to utilize unlabeled data or data prepared for other tasks to improve
the data efficiency of medical image processing, including semi-supervised learning, transfer learning
and, more recently, self-supervised learning. 
While transfer learning on large image datasets 
such as ImageNet is very common in medical image processing, 
Raghu et al. \cite{raghuTransfusionUnderstandingTransfer2019} found that
this provides little benefit in terms of model performance, but does
improve convergence speed during training.

Another way to utilize an unlabeled dataset to improve
performance of a model is semi-supervised learning, where the unlabeled
dataset is mined for soft signals to nudge the model towards better overall performance.
One such example is the one by Liu et al. \cite{liuSemisupervisedMedicalImage2020} where
they present a semi-supervised learning method for medical image classification using
a combination of unlabeled and labeled data from the same domain.

\subsubsection{Self-supervised learning}

Recently more and more papers use self-supervised learning (SSL) to pre-train neural networks on
unlabeled data, and then fine-tune the networks on the available labeled data.
SSL is a method of unsupervised neural network training where the
goal is to train an encoder that will understand useful features for a downstream
task such as object detection, classification, or similar. SSL has
been shown to improve data efficiency
\cite{chenSimpleFrameworkContrastive2020} as well as robustness to dataset imbalance
\cite{liuSelfsupervisedLearningMore2021}.

There are several approaches to training the encoder such that it learns useful features.
One approach is to use a constructed \textit{pretext} task for which one
can automatically obtain the correct solution from unlabeled data so that the correct solution can be used
for supervised training. An example of this approach is presented by Noroozi and Favaro 
\cite{norooziUnsupervisedLearningVisual2016} where the neural network is trained to 
solve a jigsaw constructed from an unlabeled dataset.

Recently, a more common approach to SSL is contrastive learning.
In contrastive learning, the encoder is trained to minimize the distance between feature
vectors of positive examples and maximize the distance between negative examples. The
positive examples are constructed in an unsupervised manner by e.g. randomly augmenting
an image twice, thus producing two examples for which the feature vectors should be
similar. Among others, notable examples of such approaches are SimCLR 
\cite{chenSimpleFrameworkContrastive2020} and MoCo \cite{he2019moco}.

\subsubsection{Self-supervised learning in medical images}

Due to its potential to improve data efficiency, SSL is a promising
approach for medical image processing. Several recent
papers proposed SSL methods both using the pretext task approach 
\cite{baiSelfSupervisedLearningCardiac2019, ZHU2020101746} as well as contrastive
learning \cite{zhouComparingLearnSurpassing2020}. Several papers evaluate self-supervised 
pretraining on an unlabeled subset of the data for medical
imaging tasks. Taleb et al. \cite{NEURIPS2020_d2dc6368} evaluate a variety of 
self-supervised pretraining approaches
for medical image segmentation in 3D MRI and CT images as well as classification on 2D 
fundus photography images. They use unlabeled data of the same modality but from a 
different corpus. Azizi et al. \cite{aziziBigSelfSupervisedModels2021} introduce a
novel contrastive learning method and evaluate it at various percentages of used
dataset labels on dermatology and chest X-ray classification. However, to 
our knowledge the are currently no papers that evaluate self-supervised 
pretraining for object detection tasks in medical images.

\section{Dataset description and demographics}\label{dataset}

The dataset used in this paper is a dataset of 15,000 labeled chest radiographs called VinDr-CXR, 
described in more detail in \cite{nguyenVinDrCXROpenDataset2021}. While the original dataset 
contains 3,000 additional test images, we were not able to obtain the labels for these images, and 
they were not used in this paper. Each scan of the dataset was labeled by three separate 
radiologists. The dataset was collected from two major Vietnamese hospitals and is to our
knowledge the largest dataset for radiograph object detection. Therefore, it is a good indicator
for the generalizability of our experiments on other radiograph datasets.

\subsection{Data preparation}

Each DICOM image from the dataset was resized to a resolution of 512 $\times$ 512 pixels. We 
discard all examples for which there is no anomaly found (examples labeled as ``no finding''). After 
discarding, we are left with a total of 4,394 images. We randomly split this dataset into a 
training set (70\%, 3,075 images), validation set (10\%, 440 images) and a test set (20\%, 878 
images). The training set was used to train the models, the validation set was used to tune the 
model hyperparameters and determine when to stop training, and the test set is used for final 
evaluation. The model did not have access to the test set during training. The class distribution
of the full dataset is shown in Fig. \ref{fig:class-balance}. During training, each image had 
a 50\% chance of being flipped horizontally.

\begin{figure}[t]
\centering
\includegraphics[width=\columnwidth]{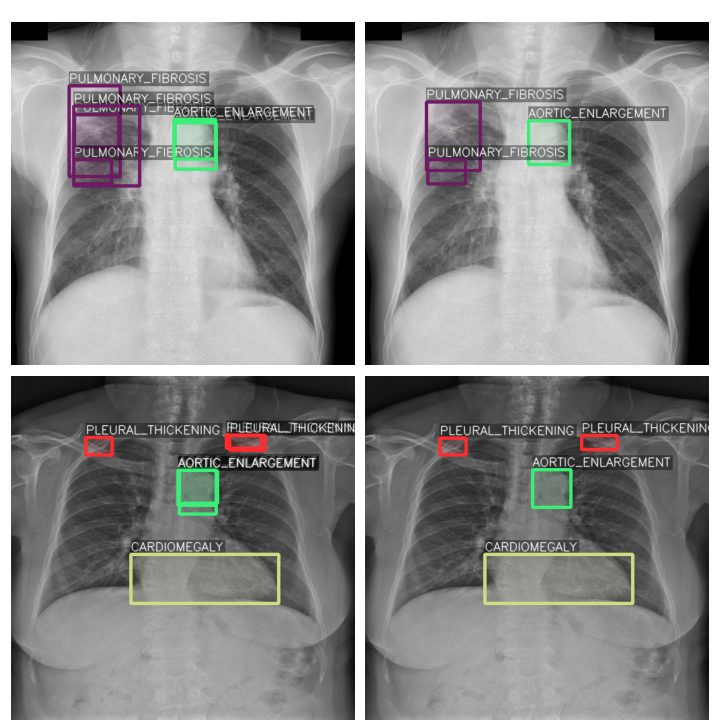}
\caption{Examples of bounding box averaging. The images on the left show the original bounding boxes, while the images on the right show fused bounding boxes which are used in our experiments.}
\label{fig:wbf}
\end{figure}

Each image can have one or more 
labels from multiple experts, and these labels often overlap. To produce the least noisy labels, 
we average overlapping bounding boxes of the same class into one bounding box. 
To determine if two boxes are overlapping, an intersection-over-union 
(IoU) threshold of 20\% is used. This approach is based on weighted boxes fusion 
described in \cite{solovyevWeightedBoxesFusion2021} but modified such that each bounding box has 
equal weight and confidence since they were manually labeled by an expert. An example of fused 
bounding boxes is shown in Fig. \ref{fig:wbf}.

\section{Methods}

\begin{figure}[t]
\centering
\includegraphics[width=\columnwidth]{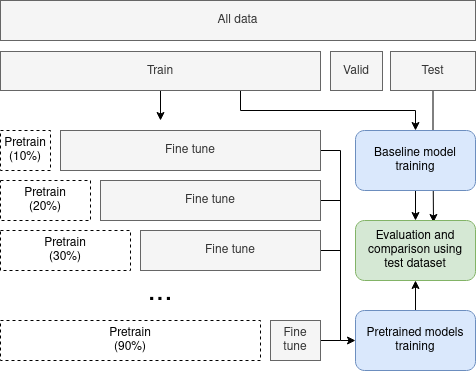}
\caption{A summary of our approach. A percentage of the training dataset is moved to the pretraining dataset and used to pretrain the model, which is then fine-tuned with the rest of the training data. The pretraining datasets are unlabeled. A separate baseline model is trained using the full labeled dataset.}
\label{fig:dataset-split-summary}
\end{figure}

A summary of our approach is shown in Fig. \ref{fig:dataset-split-summary}. 
The main goal of this paper is to analyze how self-supervised pretraining affects data efficiency for object detection 
in medical images. Therefore, we first train a \textit{baseline} deep learning model with no pretraining and on the full labeled 
training dataset following a standard approach for this type of problem, described later in this section. 
This model will be used as a point of comparison to more objectively evaluate the pre-trained models.

To evaluate the pretraining, we randomly split the training dataset into two separate datasets, a pretraining and fine-tuning 
dataset. For the pretraining dataset, we discard all class labels, as this dataset will be used to pretrain the model using 
self-supervised learning on unlabeled data. The fine-tuning dataset will then be used to fine-tune the pre-trained model using 
standard supervised learning. We train nine different pre-trained models in total, ranging from 10\% to 90\% of the total 
training set in the unlabeled pre-train dataset, in increments of 10\%.

\subsection{Balancing the dataset}

\begin{figure}[t]
\centering
\includegraphics[width=\columnwidth]{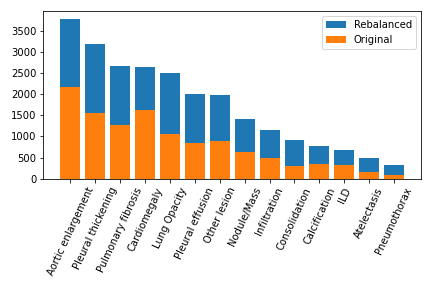}
\caption{A histogram of the class balance of the original dataset (by number of images containing the class), and the histogram of the oversampled dataset used for training.}
\label{fig:class-balance}
\end{figure}

As shown in Fig. \ref{fig:class-balance}, the dataset is highly imbalanced. To improve class balance during training we oversample examples with less-
represented classes. We use the oversampling approach described in \cite{gupta2019lvis}, with an oversampling threshold of 0.4. This balancing is performed 
online during training and only on the training set. We found that balancing the dataset in this way significantly improved the mean average precision when 
averaged across all classes.

\subsection{Baseline model details}

For an objective and fair comparison, we train a baseline model using a standard 
deep learning-based approach for object detection. We use a Faster R-CNN-based 
model \cite{DBLP:conf/nips/RenHGS15} with a ResNet-50 encoder \cite{He_2016_CVPR}. 
The baseline model is initialized using pretrained weights trained on the COCO dataset for 
12 epochs, a batch size of 2, and using the SGD optimizer with a learning rate of 
0.0002, the momentum of 0.9, and a weight decay of 0.0001. Training
is done by using the SGD optimizer with a learning rate of 0.001, a momentum of 0.9, a weight decay 
of 0.0001, and a batch size of 8. The baseline model converged in 15 epochs. We use the cross-entropy loss
for the classes and the L1 loss for the bounding boxes.

\subsection{Pretraining model details}

The pretraining model we use is a SimCLR-based model \cite{chenSimpleFrameworkContrastive2020} to pre-train a ResNet-50 backbone, the same backbone used in the baseline model. We train the SimCLR model (described later in this section) using the unlabeled pretraining dataset. We then use the pre-trained backbone in the same model as our baseline model and fine-tune the final model on the fine-tuning dataset. The result is a model similar to our backbone model but trained on fewer labeled data.

In our experiments, we use the following augmentations for SimCLR training:

\begin{enumerate}
\item A random crop and resize of the original image by a factor of 0.2 to 1.
\item A random horizontal flip (with a 50\% chance).
\item A random Gaussian blur with $\sigma$ between 0.1 and 2, and a kernel size of 21.
\item A random amount of Gaussian noise with $\sigma$ between 12.5\% and 25\% of the mean image pixel value.
\end{enumerate}

In addition, each training and validation image has histogram normalization applied.

We pretrain the models using the SGD optimizer with a learning rate of 0.001, a 
momentum of 0.9, and a weight decay of $5\cdot10^{-4}$. We also use a cosine 
annealing learning rate as described in \cite{loshchilovSGDRStochasticGradient2017}. 
For pretraining we use the NTXent loss as described in  \cite{chenSimpleFrameworkContrastive2020} with a temperature of 0.5.
The pretraining was stopped after 30 epochs in all experiments.

\section{Results}\label{sec2}

A summary of the results of our experiments is shown in Table \ref{tab:results}. All of the results in this section are calculated on the testing dataset. Our main evaluation metrics are the mean average precision (mAP), averaged across all classes and IoU thresholds from 0.5 to 0.95 in 0.05 intervals (mAP@[.5, .95]), as is standard for benchmarking the COCO dataset. We also calculate the mAP at a fixed IoU of 0.5 (mAP@0.5), which is standard for evaluating models on the PASCAL VOC dataset. In addition, we calculate the average recall given 100 detections per image (AR@100). We separately calculate the mAP[.5, .95] as well as the AR@1000 for small objects (less than 32 pixels$^2$).

\definecolor{LightGray}{gray}{0.9}

\begin{table}[t]
\renewcommand{\arraystretch}{1.3}
\caption{A summary of the results of our experiments. The details of the metrics are described in \ref{sec2}. Training images is the total number of labeled training examples available to the model.}
\label{tab:results}
\centering
\begin{tabular}{|c|c|c|c|c|c|c|c}
\hline
\begin{minipage}{5mm} ~\\ \\ \\ \end{minipage} & mAP & mAP50 & \parbox[c]{7mm}{mAP\\small} & AR & \parbox[c]{7mm}{AR\\small} & \parbox[c]{9mm}{Training\\images} \\ \hline \hline
\rowcolor{LightGray}
Baseline & 0.129 & 0.278 & 0.021 & 0.412 & 0.154 & 3075 \\ \hline
SSL 10\% & \textbf{0.142} & 0.284 & \textbf{0.026} & \textbf{0.413} & 0.156 & 2767 \\ \hline
SSL 20\% & 0.139 & \textbf{0.292} & 0.020 & 0.412 & \textbf{0.158} & 2460 \\ \hline
SSL 30\% & 0.130 & 0.268 & 0.014 & 0.402 & 0.146 & 2152 \\ \hline
SSL 40\% & 0.123 & 0.272 & 0.014 & 0.394 & 0.131 & 1845 \\ \hline
\rowcolor{LightGray}
SSL 50\% & 0.109 & 0.248 & 0.011 & 0.387 & 0.131 & 1537 \\ \hline
SSL 60\% & 0.104 & 0.230 & 0.011 & 0.378 & 0.120 & 1230 \\ \hline
SSL 70\% & 0.095 & 0.202 & 0.007 & 0.363 & 0.108 & 922 \\ \hline
SSL 80\% & 0.081 & 0.178 & 0.006 & 0.338 & 0.089 & 615 \\ \hline
SSL 90\% & 0.060 & 0.135 & 0.003 & 0.303 & 0.064 & 307 \\
\hline
\end{tabular}
\end{table}

The baseline model, trained on all of the labeled data, achieves an mAP of 0.129. In addition, by adding pretraining on a small percentage of the dataset (20\% or fewer) the performance improves for all of the metrics we calculated. For instance, when pretraining on 10\% of the data (i.e. removing 10\% of the labels) the mAP increases to 0.142. This is the best-performing model in our experiments. However, even with just 60\% of the labeled data, we achieve an mAP of 0.123, more than 95\% of the performance of the baseline model in terms of mAP. Similar results can be seen in terms of recall — the baseline model achieves an average recall of 0.412 while with 60\% of the labels we achieve an AR of 0.394, more than 95\% of the base model's AR.

However, the differences are larger when looking at small objects. While the baseline model achieves a 0.021 mAP for small objects, the model trained on 60\% of the labeled images achieves a small objects mAP of 0.014, 66.66\% of the baseline model. Similarly, the model trained on 60\% of the labeled images achieves a small object AR of 0.131, 85\% of the baseline model's small object AR of 0.154. 

\begin{figure}[t]
\centering
\includegraphics[width=\columnwidth]{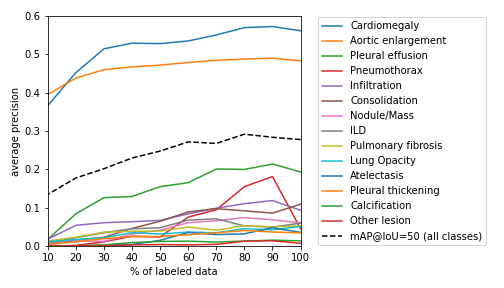}
\caption{The mean average precision (mAP@[.5, .95]) across different percentages of labeled data for different classes in the dataset. The model at 100\% of labeled data is the baseline model.}
\label{fig:classwise-map}
\end{figure}

A graphical summary of the class-wise results is shown in Fig. \ref{fig:classwise-map}. The performance on most classes drops off linearly between the baseline model and the model trained on 60\% of the labels data. The dropoff becomes more significant after training on less than 60\% of the labels. In all cases, the mAP at 60\% of the labeled data is highly correlated with the mAP across all classes as described earlier in the section, and the same conclusions apply to all classes.

The model performs the best at detecting cardiomegaly and aortic enlargement, two classes with a large number of examples and a large average object area size. 
All of the models perform significantly worse at detecting other anomalies. However, by adding pretraining on a small percentage of the data (20\% or fewer) 
the mAP improves for almost all classes, most significantly for detecting other lesions and infiltration.

We also evaluated our experiments in terms of IoU, and compared the inter-observer IoU between multiple experts and our experiments, as shown in Table 
\ref{tab:iou}. The original dataset was labeled by a group of three radiologists from a total of 17 radiologists identified as $R1$ through $R17$. In our test 
dataset, 747 images (85\% of the test dataset) were labeled by both radiologists $R9$ and $R10$, while 750 images (85.4\% of the test dataset) were labeled by 
both radiologists $R9$ and $R11$. Therefore, we calculate the inter-observer IoU between radiologists $R9$ and $R10$ as well as $R9$ and $R11$, since they 
represent the majority of the dataset. For calculating the inter-observer IoU, we only take into account labels of the same class where the IoU overlap is over 0.2, to avoid 
counting multiple instances of a class as the same prediction. We calculate the models' IoU metric as the mean of IoUs between each fused ground truth 
annotation and the maximum-overlap model prediction of the annotation's class.

\begin{table}[b]
\renewcommand{\arraystretch}{1.3}
\caption{The inter-observer IoU between radiologist $R9$ and radiologists $R10$ and $R11$, as well as the mean IoU of the model's predictions.}
\label{tab:iou}
\centering
\begin{tabular}{|c|c|c|}
\hline
& mean IoU & $\sigma$ \\ \hline
R09 vs R10 & 0.580 & 0.284 \\ \hline
R09 vs R11 & 0.674 & 0.149 \\ \hline
Baseline & 0.565 & 0.309 \\ \hline
SSL 10\% & 0.566 & 0.309 \\ \hline
SSL 20\% & 0.565 & 0.308 \\ \hline
SSL 30\% & 0.555 & 0.312 \\ \hline
SSL 40\% & 0.545 & 0.316 \\ \hline
\end{tabular}
\end{table}

The inter-observer IoU is between 0.58 and 0.67, while the IoU between ground truth and predicted 
labels of our baseline model is 0.57, slightly worse than the variability between two human 
experts. The IoU starts to drop after more than 20\% of the labels are removed. At 60\% of the 
labels, the IoU is 0.545, more than 96\% of the baseline IoU. Note that the model's IoU is based on 
the best predictions coming from the model. Under real-world conditions, it is not known which 
predictions are best and some heuristic is needed which might not always choose the best 
prediction, so the presented IoU values should be interpreted as best-case scenario IoU values. 
Still, the relative IoU differences between the baseline and the pretrained models should remain the same.

\section{Discussion and conclusion}\label{sec12}

Our results on medical object detection experiments show that it is possible to use fewer labeled 
examples and still achieve similar performance by utilizing self-supervised 
pretraining. All of the global metrics we measured, including average precision and recall as well 
as IoU, remain within 95\% of the baseline model's performance. The same is true for class-based 
metrics. In our experiments, using 60\% of labeled data is a good compromise between using fewer 
labels while still maintaining good object detection performance. We hypothesize that the 
backbone trained under self-supervised contrastive learning produces more salient feature maps and 
thus overcomes the lack of labels when compared to the baseline model. 

The biggest performance loss for the self-supervised models is in smaller objects, where the model 
trained on 60\% of the labeled data achieves less than 65\% of the small object mean average 
precision of the baseline model. A very common occurrence in deep learning models for 
object detection is a gap in performance between predicting small and large objects. Smaller objects 
are less frequent, and cover a smaller area of the image, making it difficult for models to predict 
them. This problem can be overcome at least in part using augmentation 
\cite{kisantalAugmentationSmallObject2019}, and it's possible our results could be improved by 
adding oversampling and small object augmentation to both the training and pretraining steps.

However, at 60\% of labeled data, the model still retains 85\% of the small object recall of the baseline model.
In medical image processing tasks, recall is often more important than precision as the cost of finding false positives
is much lower than the cost of missing true positives. Therefore, the reduction in small object detection performance
is less severe than it appears from the mAP.

We hope that the findings of this paper will lead to enabling more deep learning-based methods in 
medical object detection with fewer labeled data, as well as increase the performance of existing 
models trained on small datasets. This paper is also an indicator that self-supervised learning is 
a very promising avenue for future research in medical image processing.

\bibliography{bibfile.bib}

\end{document}